\pdfoutput=1

\documentclass[11pt]{article}

\usepackage[]{acl}

\usepackage{times}
\usepackage{latexsym}

\usepackage{graphicx}
\usepackage{subcaption,booktabs}
\usepackage{tabularx}
\usepackage{multirow, amsmath}
\usepackage{placeins}
\usepackage{gensymb} 
\usepackage{amssymb}
\usepackage{setspace}
\usepackage[T1]{fontenc}

\usepackage[utf8]{inputenc}

\usepackage{microtype}

 \usepackage{indentfirst}
\usepackage{natbib}
\usepackage{csquotes}
\usepackage{amsmath}
\usepackage{amssymb}
\usepackage{booktabs}
\usepackage{latexsym}

\usepackage{url}
\usepackage{booktabs}
\usepackage{multirow}
\usepackage{array, caption, floatrow, makecell, booktabs}
\usepackage{wrapfig}
\usepackage{floatrow}
\usepackage{comment}
\usepackage{enumitem}
\usepackage{accents}
\usepackage{xspace,mfirstuc,tabulary}
\usepackage{tabu}
\usepackage{xcolor}

\usepackage{bm}
\usepackage{todonotes}

\newcommand{\our}{DisinfoMeme\xspace}

\title{DisinfoMeme: A Multimodal Dataset for Detecting Meme Intentionally Spreading Out Disinformation}
\author{Jingnong Qu$^{1}$, Liunian Harold Li$^{1}$, Jieyu Zhao$^{2}$, Sunipa Dev$^{3}$, Kai-Wei Chang$^{1}$ \\
$^1$University of California, Los Angeles \quad $^2$University of Maryland, College Park \\
$^3$Google Research\\
\texttt{\small{andrewqu2000@ucla.edu, liunian.harold.li@cs.ucla.edu,}} \\
\texttt{\small{jieyuz@umd.edu, sunipadev@google.com, kwchang@cs.ucla.edu}} 
}

\date{April 2022}

\begin{document}

\maketitle
\begin{abstract}
Disinformation has become a serious problem on social media. In particular, given their short format, visual attraction, and humorous nature, memes have a significant advantage in dissemination among online communities, making them an effective vehicle for the spread of disinformation. We present DisinfoMeme to help detect disinformation memes. The dataset contains memes mined from Reddit covering three current topics: the COVID-19 pandemic, the Black Lives Matter movement, and veganism/vegetarianism. The dataset poses multiple unique challenges: limited data and label imbalance, reliance on external knowledge, multimodal reasoning, layout dependency, and noise from OCR. We test multiple widely-used unimodal and multimodal models on this dataset. The experiments show that the room for improvement is still huge for current models.
\end{abstract}

\begin{figure*}
\includegraphics[width=\textwidth]{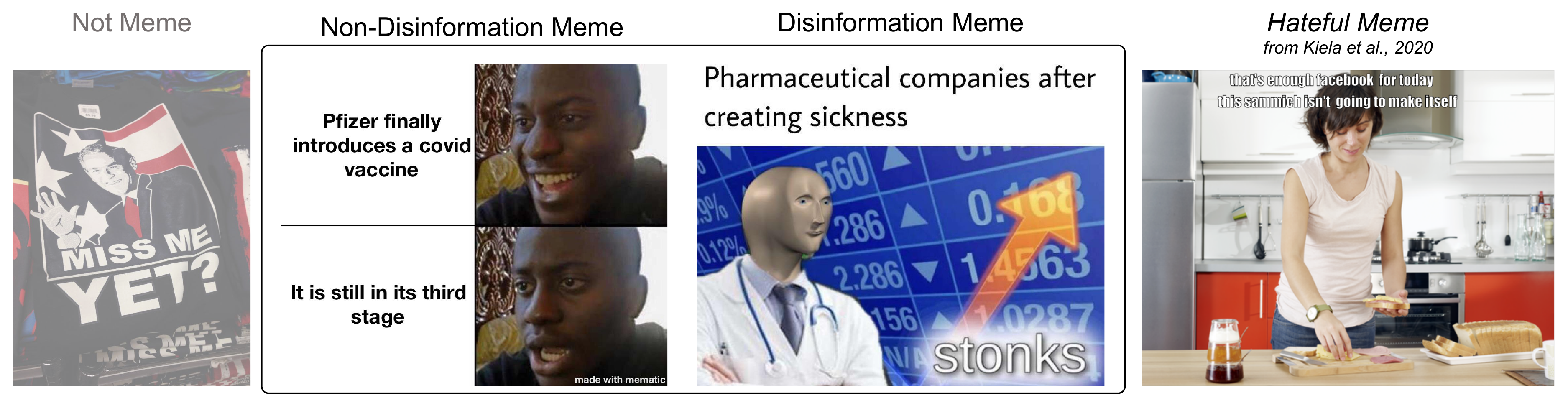}
\caption{\label{lead}
Comparison between a non-meme image and three kinds of memes. Memes contain human efforts to combine text and images as illustrated by the three memes here. Although some images contain text, it does not automatically qualify them as memes. Memes in \our differ from those in Hateful Memes \citet{NEURIPS2020_1b84c4ce} in layout, style, and content.
}
\end{figure*}
\section{Introduction}
\label{sec:intro}

\noindent
The development of social media has greatly facilitated the spread of information. \citet{shearer2017news} report that around two-thirds of US adults use social media to get news. However, the spontaneity of social media also invites the rapid spread of inaccurate information. Therefore, misinformation/disinformation has become a serious problem on social media. To give an example, during the COVID-19 pandemic, while social media allows healthcare practitioners to communicate professional information to the public promptly \citep{Gottlieb2020}, previous studies also show that the spreading of inaccurate information about public health could cause severe consequences \cite{broniatowski2018weaponized,swire2019public,siddiqui2020social,ferrara2020misinformation}.

Of particular interest, is the prevalence of memes as vehicles for the spread of inaccurate information. A meme is an image overlayed with text and created by Internet users; we define ``disinformation memes'' as those designed to actively spread inaccurate information. The third image in Figure~\ref{lead} shows a disinformation meme that attempts to spread the idea that pharmaceutical companies acquire financial gains through inventing diseases. We are concerned with disinformation memes as it has been argued that memes can serve as more effective tools for the spread of inaccurate information than pure text \cite{zannettou2018origins,hameleers2020picture,li2020picture,ferrara2020misinformation}. In contrast, prior work on disinformation detection has largely focused on fake news detection, and few works utilize images for their purpose \cite{thorne2018fever,zellers2019defending,bozarth2020toward,alam2021survey}.

In this work, we construct \our, a multimodal dataset for detecting memes intentionally designed to spread out disinformation. \our contains 1,170 annotated memes mined from Reddit related to three representative social topics -- the COVID-19 pandemic (\textbf{Pandemic}), the Black Lives Matter movement (\textbf{BLM}), and Veganism/Vegetarianism (\textbf{Veganism}). \our also contains 30,302 \textit{unannotated} images to encourage the development of unsupervised or weakly-supervised learning methods~\cite{li2020unsupervised}. 

\our presents several unique challenges (Section \ref{sec:3_challenge}). 1) \textit{Limited data and label imbalance}. Disinformation memes are the minority of Internet images as most images in online forums are not memes, and most memes are not designed to be disinformation. Our paper addresses the inherent imbalance by proposing to conduct image retrieval only in meme-prone communities based on our disinformation-prone keywords, so we could construct a feasible problem that considers that real-world scenario. 2) \textit{The reliance on external knowledge}. Memes refer extensively to popular cultures, slang, and current events. For example, understanding the third meme in Figure~\ref{lead} requires the understanding of the word `stonks', which means gaining substantial profits financially\footnote{\url{https://www.urbandictionary.com/define.php?term=Stonks}}. Thus, detecting disinformation memes requires the model to be well-versed in such domain-specific knowledge. 3) \textit{Multimodal reasoning}. As a combination of text and image, the understanding of many memes requires taking both modalities into consideration. 4) \textit{Layout dependency}. A key distinction between meme understanding and generic vision-and-language tasks (e.g., VQA~\cite{antol2015vqa}) is that many memes can be viewed as structured documents~\cite{ha1995recursive,xu2020layoutlm} and the layout (the relative position of text and images) is crucial. Changing the layout might break the semantic meaning of the meme (see the second example in Figure \ref{lead}). 5) \textit{OCR noise}. Internet memes leverage a wide range of images and text of different styles; they also have a great variety in layouts. This could create great difficulty for OCR systems to accurately extract the text from memes.

We test several widely-used unimodal and multimodal models on \our (Section \ref{sec:4_exp}). Results show that while the models can achieve non-trivial results, the performance is still far from satisfactory; in addition, current multimodal models only marginally or even fail to outperform unimodal models, suggesting they are not successful at utilizing information from both modalities for detecting disinformation memes. We conclude that the dataset constitutes a meaningful and challenging task for vision-and-language models and we will release the dataset to facilitate future research.

\section{Dataset}
The dataset utilizes user content from Reddit. Reddit is a social media platform that hosts communities that are divided based on topics. These communities are called \textit{subreddits}. 
Users can make \textit{posts} and \textit{comment} them in each subreddit. 
To build this dataset, we retrieve images that are potentially topic-related memes from posts in selected subreddits and assign the annotation tasks for the images on Amazon Mechanical Turk (MTurk). In the annotation tasks, we asked the MTurk workers to decide whether an image is a topic-related meme and whether it promotes disinformation. Reddit grants API users a non-transferable right to copy and display User Content from Reddit in accordance with the format of display\footnote{\url{https://www.reddit.com/wiki/api-terms}}, we comply with the relevant terms imposed by Reddit.

\subsection{Definitions}
\label{sec:3_1_definitions}
To annotate disinformation memes, we must define what is a \textit{meme} and what constitutes a \textit{disinformation meme}. We define ``meme'' as follows:

\begin{displayquote}
An image that contains images along with text pieces that are artificially superimposed in or around the images, which need to be comprehended together with the text in order to get the full meaning. 
\end{displayquote}

This definition matches the intuitive appearance of an Internet meme and enforces a requirement of multimodal understanding on top of that. 
We also make a distinction between misinformation and disinformation. \citet{wu2019misinformation} use the term ``misinformation'' to refer to ``all false or inaccurate information that is spread in social media.'' The term ``disinformation'', on the other hand, indicates an intention to spread inaccurate information. 

In this study, we choose to focus on disinformation. Due to the humorous nature of many memes, the intention of memes could be satirical in the content it displays. A previous study also showed that memes have been used in political satire \citep{kulkarni2017internet}. To promote healthy social media discussion, we believe it would be vital to distinguish instances that are not promoting misinformation and even criticizing it from actual disinformation. 

Judging whether a piece of text is disinformation often involves an inference task in semantics or pragmatics. For example, the inaccurate information in the disinformation meme in Figure~\ref{lead} is never explicitly stated but formed through an inference. Since disagreement is inherent in the annotation of natural language inference tasks~\cite{pavlick2019inherent}, we would expect annotator disagreement to a degree. We take a utilitarian view towards this issue: for disinformation meme detection, the gradient nature of the problem should be taken into consideration. Therefore, annotators' choice is transformed to a score of $1$ as not actively spreading inaccuracy information, $2$ as hard to tell, and $3$ as actively spreading inaccuracy information. Then, we calculate the mean score of 3 workers for the same image. 
It is important to catch most cases of disinformation in this detection task, so recall is the more important metric than precision as it concerns false negatives. Thus, we regard examples with an annotator score larger than 2 (a moderately low value) as disinformation memes (positive examples) and other examples as non-disinformation memes (negative examples). 

\begin{figure*}
\includegraphics[width=\textwidth]{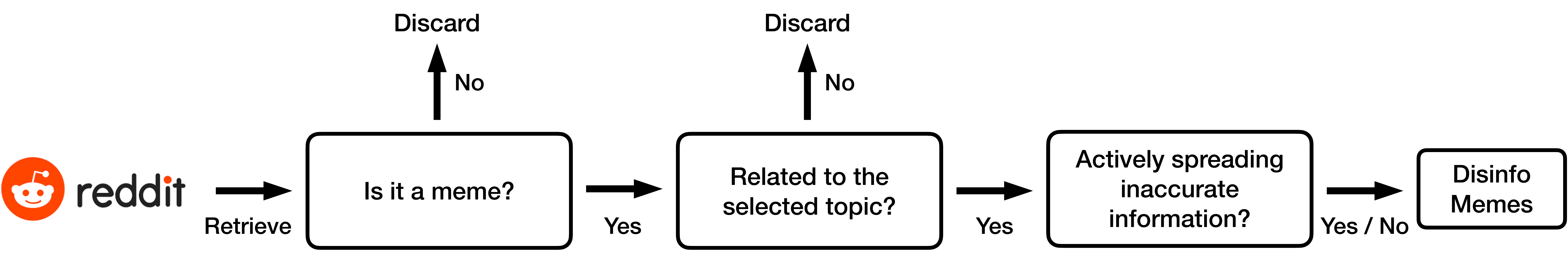}
\caption{\label{flow}
Images are first retrieved from Reddit. Then they go through the annotation process, where a negative answer in the first two questions would lead to the discard of the image. Refer to Figures~\ref{Q1},~\ref{Q2}, and~\ref{Q3} in the appendices for interfaces of annotation assignments.
}
\end{figure*}
\subsection{Dataset Construction}

The dataset contains a supervised set and an unsupervised set, both spanning three topics of Pandemic, BLM, and Veganism. The first two topics consist of heated debate and easily disseminated disinformation that could lead to severe consequences and divided society, while the third serves as a contrast for being a less controversial topic. More topics could be added in the future following the annotation process. The supervised set was sampled from the unsupervised set and underwent human annotation by MTurk workers. 

\paragraph{Retrieve images from Reddit.}

The dataset aims to include the most up-to-date contents by the time of collection as the topics are tightly related to current events. Therefore, we opted to use PRAW\footnote{\url{https://github.com/praw-dev/praw}}, a wrapper of Reddit API, to retrieve images in real-time manner from Reddit. We first decided on a group of topic-related keywords. Then, we used these keywords to find more related words using Related Words\footnote{\url{https://relatedwords.org/}}. After these procedures, we made additional selections and additions for the words to ensure the relevantness and completeness. 

In addition, we also wanted to make sure that the queries would be likely to return an image that was a meme. Therefore, we restricted queries to subreddits where users are likely to post memes. The process of selecting subreddits was analogous to that of selecting keywords. We started with several famous subreddits related to memes across the political spectrum. Then, we used these subreddits to find related subreddits using Related Subreddits\footnote{\url{https://anvaka.github.io/sayit/}}. Human inspection of individual subreddits was involved to select those where members were actively posting memes in their posts. Then queries on the keywords were conducted in each of the selected subreddit. For each post with an image in the post, we will retrieve the image and record relevant information about the post. The text in the images was extracted with the OCR of the Google Cloud Vision API\footnote{\url{https://cloud.google.com/vision/docs/ocr}}.

\paragraph{Annotation process.}

After receiving the outcome of a pilot study of the annotation process, we designed an annotation process that involves three steps. In each assignment of the first step, we first showed the MTurk workers our definition of meme. Then, they will be shown 10 images. For each image, they will be asked whether it is a meme. Each assignment will be viewed by two different workers. Only the images where both workers agree that they memes will enter the next step. In the second step, the memes are shown in a similar structure to the workers, and the workers are asked to determine whether the meme is related to one of the three topics, depending on which topic the meme is selected from. Three workers will work on each assignment. Only the memes where at least two workers agree that the meme is related to the selected topic will enter the final step. In the final step, the workers will be asked to determine whether the meme shown here is actively spreading inaccurate information based on their perception. As noted in Section \ref{sec:3_1_definitions}, we provide the annotators with 3 choices and calculate an average score among them after transforming each choice into a score ranging from 1 to 3.

\subsection{Dataset Statistics}
\label{sec:data_stats}

We present \our, a dataset for detecting memes intentionally spreading out disinformation. \our contains 30,302 images in the unsupervised set and 1,170 of them in the supervised set. Table~\ref{tab:topic-num} introduces the basic statistics of the supervised set with 980 memes in the Pandemic topic, 74 memes in the BLM topic, and 116 memes in the Veganism topic. We provide a positive and a negative example in each of the three topics in Figure~\ref{more} These 1,170 memes are annotated from a set of 2,893 images. 926 of them were eliminated for not being a meme, and the rest 797 were eliminated for not being topic-related.

\section{Challenges in \our}
\label{sec:3_challenge}
We examine some statistical and qualitative characteristics of \our dataset. These characteristics pose unique challenges to the problem of real-world multimodal disinformation detection.
\begin{figure*}[t]
\includegraphics[  width=\textwidth]{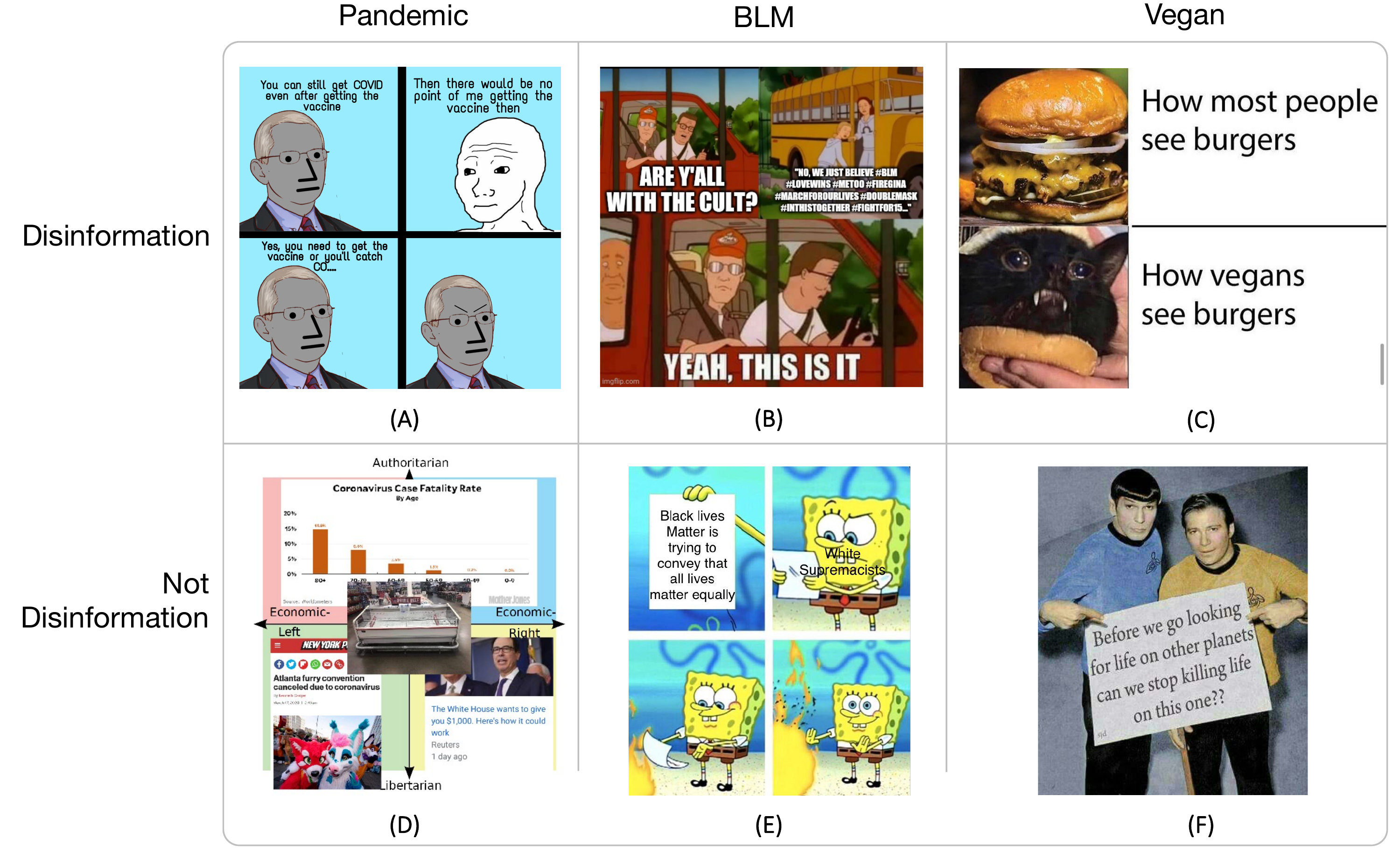}
\caption{\label{more}
Example memes from the three topics in \our. In addition to the great variety of layouts, understanding memes in \our is challenging because of the reliance on external knowledge, multimodal reasoning, the referential relationship between text and specific portions of the image, and OCR noises.
}
\end{figure*}

\begin{center}

\begin{table}
\resizebox{\linewidth}{!}{
    \begin{tabular}{l|c|c}
    \toprule
     Topic &  Disinformation & Non-Disinformation \\
     \midrule
    Pandemic & 218 & 762 \\
    BLM & 31 & 43 \\
    Veganism & 31 & 85 \\
     \midrule
    \textbf{Total} & \textbf{280} & \textbf{890} \\
    \bottomrule
\end{tabular}}
\caption{\label{tab:topic-num}Breakdown of number of disinformation memes and non-disinformation memes in each topic (the supervised set). The dataset shows significant imbalance between labels, which aligns with the real-world scenario.}
\end{table}
\end{center}

\paragraph{Limited data and label imbalance.} The first and foremost challenge in trying to build a system to combat disinformation memes is the lack of data. When building \our, we notice that disinformation memes are \textit{inherently} hard to find. 

The majority of memes on the Internet are benign and serve as normal ways of communication. This causes a problem for the dataset construction process. If we blindly collect images from unconstrained Internet forums, most images would either not be a meme or not spread disinformation. Thus, to obtain an adequate number of disinformation memes, one would need to annotate a much larger number of images, adding to the annotation cost. In this paper, we provide one potential solution: we first determine a set of communities where a lot of memes are potentially posted, and then we retrieve images only in these communities based on our selected keywords of disinformation-prone topics. Even with such a procedure, the final dataset is still comparatively small and imbalanced in label distribution as shown in Table~\ref{tab:topic-num} and Section~\ref{sec:data_stats}. Out of 2,893 images we crawled, only less than half of them are topic-related memes (1,170) and less than $1/4$ are disinformation memes (280). 

Notice that the lack of data and label imbalance is inherent to disinformation memes. Thus we hope \our could inspire future work to explore unsupervised methods \cite{li2020unsupervised} and learning with unbalanced data.

\paragraph{External knowledge.} Internet memes rely extensively on references to popular cultures, slang, current events, and domain-specific knowledge. For example, Meme (D) in Figure~\ref{more} refers to multiple news events while Meme (B), (E), and (F) all feature characters in popular TV shows. We manually go over 50 examples in our dataset and check if they require external knowledge. Results suggest that 70\% of the examples in the dataset require external knowledge (Table \ref{table:challenge}). 

\paragraph{Multimodal reasoning.}
Understanding most of the memes in this dataset requires multimodal reasoning (vision and language), as the sampling in Table~\ref{table:challenge} shows. Meme (C) in Figure~\ref{more} is one such case. To understand this meme, one has to make the connection between the texts and the images. Taking either the text or the image out would render the meme meaningless.

\paragraph{Layout dependency.}
In some memes, the semantic meaning is dependent on its layout (i.e., the referential relationship between a portion of the text to a subsection of the image). Many of the memes in Figure~\ref{more} show this property. For example, Meme (A) represents a dialogue where each paragraph of the text belongs to one of the two characters in the image. 
Meme (E) represents a more complicated relationship, where the phrase ``white supremacists'' refers to a cartoon character who burns the paper that he holds. In this sense, memes can also be viewed as structured documents~\cite{ha1995recursive,xu2020layoutlm} and require structured modeling.

\paragraph{OCR noise.}
Meme images tend to pose a great challenge for OCR systems. We used Google Cloud Vision API for OCR as an initial step. We demonstrate two examples of the OCR transcription for Figure \ref{more} below. For Meme (A), the OCR result is satisfying because of its simple layout. However, for Meme (D), the result is messy as text is covered by images and next to each other. Another challenge is the watermark (e.g., the second meme in Figure \ref{lead} has a visible watermark). OCR systems might include it in the transcription as part of the normal text and disrupt the original content.

\noindent\fbox{%
    \parbox{0.45\textwidth}{
    \begin{footnotesize}
    \begin{singlespace}
\textit{OCR for Meme (A) in Figure \ref{more} :} You can still get COVID even after getting the vaccine. Then there would be no point of me getting the vaccine then. Yes, you need to get the vaccine or you'll catch CO...

\textit{OCR for Meme (D) in Figure \ref{more} :} Authoritarian Coronavirus Case Fatality Rate By Age 20\% 11.. 15 10\% G.0 1.5\% 12.h 0.0 0\% 20-70 50-59 40-19 0-9 Mother Jomes Economic- Source. Morldameters PRIME BEEF Economic- Left NEW YORK P Right Atlanta furry convention canceled due to coronavirus The White House wants to give you \$1,000. Here's how it could work Reuters 1 day ago Libertarian
\end{singlespace}
\end{footnotesize}

}
}

\paragraph{Comparison with Hateful Memes.}
\our is inspired by Hateful Memes~\cite{NEURIPS2020_1b84c4ce} but poses a unique challenge as detailed above.
Here, we sample 50 examples from \our and another 50 examples from the Hateful Memes dataset for comparison. We list the number of memes that satisfy each of the five challenges in each dataset in Table~\ref{table:challenge}. Note that the Hateful Memes dataset is constructed in a controlled experimental setting: the meme texts are checked by annotators and then overlaid onto a clean image; thus OCR is not used in the dataset. In contrast, we operate in a more realistic albeit noisy setting where the memes are directly mined from the Internet and require OCR.
Figure~\ref{lead} also provides an example from \citet{NEURIPS2020_1b84c4ce} to show that memes in that dataset tend to have simpler layouts than memes in \our. 

\begin{table}
\centering
\resizebox{\linewidth}{!}{
\begin{tabular}{lll}
\toprule
\textbf{Challenge}&\textbf{\our}&\textbf{Hateful Memes}\\
\midrule
External knowledge & 70\% & 60\%\\
Multimodal reasoning & 74\% & 52\%\\
Layout dependency & 46\% & 6\%\\
OCR Noise & 68\% & -- \\
\bottomrule
\end{tabular}}
\caption{\label{table:challenge}
Breakdown of the challenge in \our and Hateful Memes \citep{NEURIPS2020_1b84c4ce} on samples of 50 examples. \our poses many new challenges to multimodal learning.
}
\end{table}

\section{Experiments}
\label{sec:4_exp}
\begin{table}
\centering
\resizebox{\linewidth}{!}{
\begin{tabular}{ll|cccc}
\toprule
& \multirow{2}{*}{\textbf{Model}} & \multicolumn{4}{c}{\textbf{Pandemic}} \\
& & ROC AUC & F1 & Precision & Recall  \\
\midrule
\multirow{2}{*}{Trivial} & Random Guess & 49.4 & 21.0 & 20.9 & 21.2  \\
& Minority Vote &50.0& 35.8& 21.8& 100.0 \\
\midrule
\multirow{4}{*}{Unimodal} & Text BERT&62.0 & \textbf{37.8} & 28.3&\textbf{58.3}  \\
& Image Grid& 53.6 &27.9& 23.3& 35.2 \\
& Image Region&56.8 & 32.3 & 25.5& 44.6 \\
\midrule
\multirow{4}{*}{Multimodal} &VisualBERT COCO&63.1 & 37.3 &29.3&55.1 \\
& VisualBERT HatefulMemes&55.7 & 33.3 &25.0&51.5\\
&ViLBERT CC&\textbf{63.2} &36.4 &\textbf{29.6}&48.6 \\
& ViLBERT HatefulMemes&59.7 &34.5 &28.3&47.8 \\
\bottomrule
\end{tabular}}
\caption{\label{test}
Experiment results on the \textbf{Pandemic} split. Models achieve non-trivial performance compared to simple baselines.
}
\end{table}
In this section, we test several widely-used unimodal and multimodal models on \our as an initial step towards detecting disinformation memes. Results suggest that 1) models can achieve non-trivial results; 2) current multimodal models marginally outperform unimoal models; 4) current models cannot utilize knowledge from the Hateful Memes dataset to better identify disinformation memes; 5) limited cross-topic transfer is possible.

\paragraph{Setup}

All the experiments are conducted on the supervised dataset. We use only the \textbf{Pandemic} category for training.  We used the Pandemic category to make 5 pseudo-random 6:2:2 train-val-test splits to counter the variance brought by the small size of the supervised set. Data under the other two topics are used as a transfer learning task for evaluating all the models. We then report the mean F1 scores, precision scores, and the area under the
receiver operating characteristic curve (ROC AUC) (\citealt{bradley1997use}; \citealt{NEURIPS2020_1b84c4ce}) of the checkpoints trained on the 5 splits for each model in all experiments. We run the experiments under the MMF \citep{singh2020mmf} framework. 

We tested on 3 unimodal models and 2 multimodal VLP models with 2 checkpoints each. The unimodal models include a text unimodal model and two image unimodal models from \citet{NEURIPS2020_1b84c4ce}, where the text unimodal model is BERT \citet{devlin2018bert} (\textbf{Text BERT}), and the two image unimodal models are standard ResNet-152 \citep{he2016deep} convolutional features from res-5c
with average pooling (\textbf{Image Grid}) and features from fc6 layer of Faster-RCNN \citep{ren2015faster} with ResNeXt 152 as its backbone \citep{xie2017aggregated}. We also include a (\textbf{Image Region}) model based on an object detection pre-trained on Visual Genome~\cite{Krishna2017visualgenome,singh2018pythia}.
For the multimodal models, we utilized VisualBERT \citep{li2019visualbert} pretrained on COCO (\textbf{VisualBERT COCO}) \citep{chen2015microsoft} and Hateful Memes followed by COCO (\textbf{VisualBERT HatefulMemes}),  and ViLBERT \citep{lu2019vilbert} pretrained on Conceptual Captions (\textbf{ViLBERT CC}) \citep{sharma2018conceptual} and Hateful Memes followed by Conceptual Captions (\textbf{ViLBERT HatefulMemes}) from \citet{singh2020mmf}.

We report two simple baselines: \textbf{Random Guess} and \textbf{Minority Vote}. The Random Guess baseline includes 5 predictions based on the distribution of the 5 validation sets respectively. The prediction for Minority predicts all examples to be positive (disinformation). In this case, the recall score will always be 100, and the precision will be non-zero, leading to a relatively high F1 score.
\begin{figure}
\includegraphics[width=\textwidth]{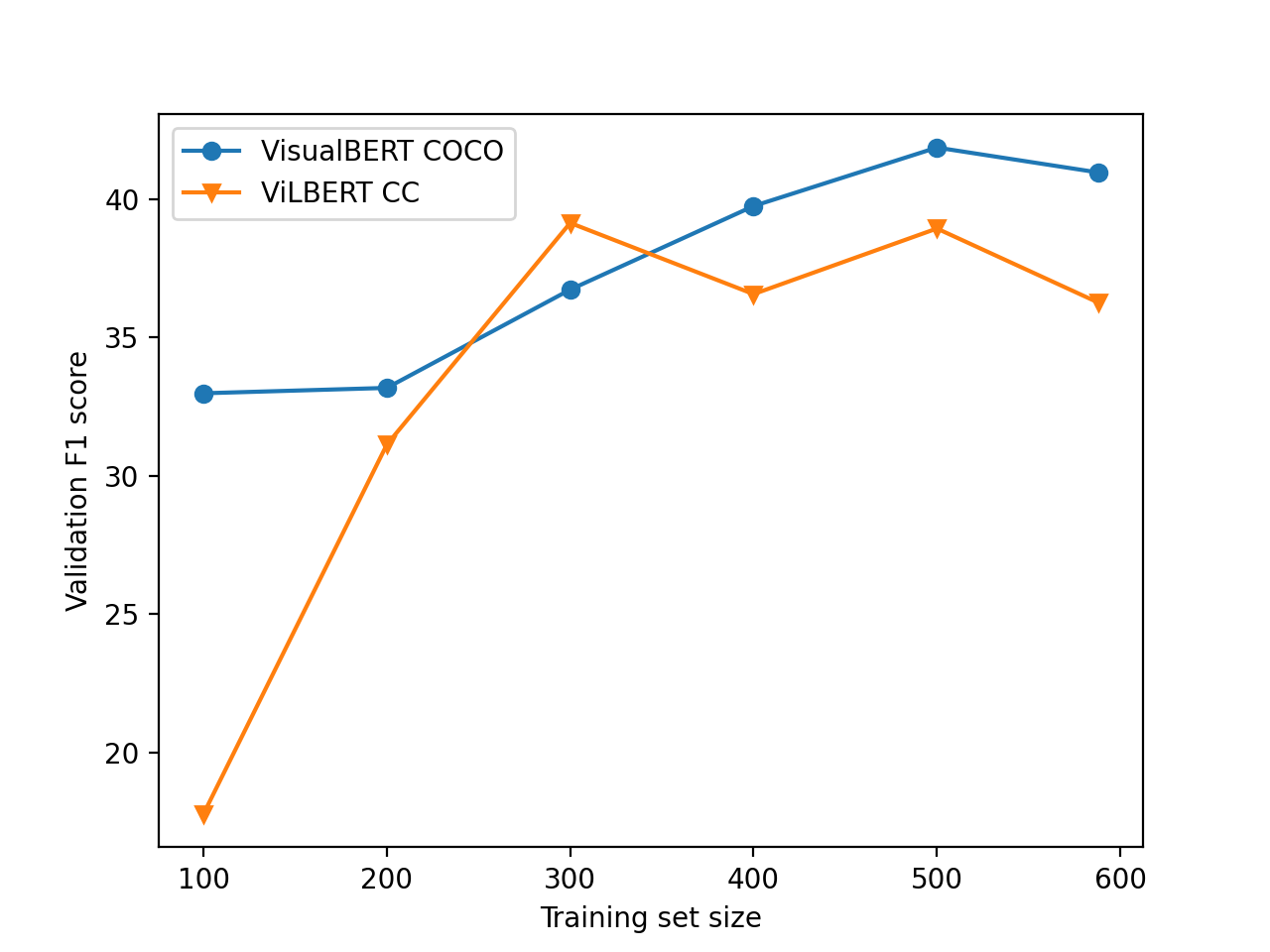}
\caption{\label{lc}
Performance of VisualBERT and ViLBERT with respect to the size of the training data. The consistent upward trend demonstrates models could benefit from more annotated data.}
\end{figure}
\paragraph{Results.} Experiments are presented in Tables \ref{test} and~\ref{table:transfersetting}. Generally, models can achieve higher performance on the recall score than the precision score. The recall score being higher is a better scenario than the opposite scenario as we would care more about the number of false negatives in combating disinformation. 

\begin{table*}
\centering
\resizebox{0.9\linewidth}{!}{
\begin{tabular}{ll|cccc|cccc}
\toprule
& \multirow{2}{*}{\textbf{Model}} & \multicolumn{4}{c}{\textbf{BLM}} & \multicolumn{4}{c}{\textbf{Veganism}} \\
&  & ROC AUC & F1 & Precision & Recall  & ROC AUC & F1 & Precision & Recall\\
\midrule
\multirow{2}{*}{Trivial} & Random Guess  & 49.6 & 28.6 & 41.8 & 21.9 & 51.3 & 25.4 &  29.4 & 22.6 \\
 & Minority Vote  & 50.0 & 59.0 & 41.9 & 100.0 & 50.0 & 42.2 & 26.7 & 100.0 \\
\midrule
\multirow{4}{*}{Unimodal} & Text BERT& 52.5 & \textbf{57.0} & 45.9& \textbf{75.5}& 43.1& 30.8 &22.9 &\textbf{47.7}   \\

& Image Grid &52.3&43.1& 47.8& 40.0&46.0 &22.6& 24.1& 21.3  \\
& Image Region &\textbf{58.1}&51.6 & 48.0& 56.1& 55.1&30.0& 29.3& 31.0 \\
\midrule
\multirow{4}{*}{Multimodal} &VisualBERT COCO &55.1& 53.3 &44.3&68.4&48.8& 30.6 &27.4&37.4 \\
& VisualBERT HatefulMemes&50.6 & 46.2 &42.4&51.6&46.5& 31.1 &26.9&38.1\\
&ViLBERT CC &52.1& 43.6 &44.9&43.2&\textbf{55.7}& \textbf{32.8} &\textbf{30.5}&39.4 \\
& ViLBERT HatefulMemes &56.6& 51.4 &\textbf{48.7}&57.4&52.8& 32.8  &29.5&40.6 \\
\bottomrule
\end{tabular}}
\caption{\label{table:transfersetting}
Experiment results in the transfer learning setting. All models are trained with the training sets from the \textbf{Pandemic} split and tested on the \textbf{BLM} and \textbf{Veganism} split.
}
\end{table*}

\paragraph{Can multimodality help?}
To understand whether the multimodal models help with the problem, we compare the VisualBERT COCO and ViLBERT CC with the three unimodal models in Table~\ref{test}. We note that the two multimodal models show similar performance on the test set F1 score as Text BERT and slightly more superior performance on ROC AUC. Along with the analysis in Section~\ref{sec:3_challenge}, we conclude that multimodal information could benefit the task but the room for improvement is still large.

\paragraph{Data scalability.}
Figure~\ref{lc} shows the learning curve of the VisualBERT COCO and ViLBERT CC. The clear upward trend in both models shows that the dataset poses a reasonable problem for multimodal models and more gains could come from more annotated data. 

\paragraph{Transfer from Hateful Memes.}
To investigate whether prior knowledge from Hateful Memes would benefit the task, we focus on VisualBERT HatefulMemes and ViLBERT HatefulMemes in Table~\ref{test}. They both show a slight performance degradation from VisualBERT COCO and ViLBERT CC respectively. This result corroborates with our observation in Section~\ref{sec:3_challenge} that DisinfoMeme poses a different problem from Hateful Memes \citep{NEURIPS2020_1b84c4ce} and current models cannot utilize knowledge from the Hateful Memes dataset well for this task.

\paragraph{Transfer across topics.}
We show the performance of all models tested with the examples under the BLM and Veganism topics in Table~\ref{table:transfersetting}. None of the models has seen any of the examples under these two topics during training.
Results show that on the two topics, the models only outperform the random guess baseline marginally. The results are reasonable as disinformation meme detection could be highly topic-dependent as different topics involve different background knowledge. Therefore, we conclude that transfer learning across topics is possible but limited.

\section{Related Work}
\paragraph{Detecting disinformation.} 
A long line of research has been devoted to detecting disinformation on the Internet. Most work focuses on fake new detection in the text domain~\cite{perez2017automatic,thorne2018fever,zellers2019defending} while some recent work extends to other modalities such as tables~\cite{chen2019tabfact} and images~\cite{wang2018eann,tan2020detecting,fung2021infosurgeon}. Our study focuses on a specific yet widely-spread form of disinformation (memes) and features unique challenges not seen in prior work (Section \ref{sec:3_challenge}) . 

\paragraph{Vision-and-Language tasks.} Memes involve the interplay between vision and language; thus \our naturally is a vision-and-language task. There has been a long line of work designed to test multimodal intelligence, such as visual question answering~\cite{antol2015vqa,Goyal2017vqa2}, image captioning~\cite{lin2014microsoft}, and visual commonsense reasoning~\cite{zellers2019recognition}. Some of the challenges in \our are also noted in prior work, such as grounding to external knowledge~\cite{marino2019ok} and understanding text in visual scenes \cite{mishra2019ocr}. Closest to our work is Hateful Memes~\cite{NEURIPS2020_1b84c4ce}, which is designed to detect hate speech on the Internet. However, Hateful Memes target unconstrained hateful meme detection, while \our features a focused study on disinformation memes about three carefully-chosen topics (see Section \ref{sec:3_challenge} for comparison with Hateful Memes).

\paragraph{Vision-and-Language models.}
There is a long line of work seeking to build a model to understand the visual world and how it is expressed in natural language~\cite{kulkarni2013babytalk,anderson2018bottom,lu2019vilbert,tan2019lxmert,chen2019uniter,su2019vl,li2019visualbert,li2019unicoder,zhou2019unified,li2020oscar,li2020unsupervised,zhang2021vinvl,radford2021learning,li2021grounded,alayrac2022flamingo}. In this study, we have tested several widely-used V\&L models but they still struggle to solve the task well. \our raises several unique challenges and we hope it could encourage the development of new V\&L methods (e.g., knowledge-enriched V\&L models \cite{zhu2020mucko,shen2022k}, unsupervised/weakly-supervised learning \cite{li2020unsupervised,zhou2022unsupervised}, and models that could understand text in images \cite{singh2019towards}).

\section{Conclusion}
In this paper, we present a dataset for multimodal disinformation detection in Internet memes. We have demonstrated that this dataset could serve as a concrete basis for the detection of multimodal disinformation on social media. We hope that \our could inspire future research on new methods for multimodal disinformation detection. 

\section*{Ethical Considerations}
We propose \our in hope that it serves as a starting point for developing automatic methods for identifying misinformation memes. We note that the dataset and subsequently the models trained with it could contain biases~\cite{zhao2017men,zhao2019gender} and other ethical issues. A disinformation meme detection model should not be deployed in the real world without a comprehensive ethical review or human moderation.

\section*{Acknowledgement}
This work is supported in part by an Amazon Fellowship.

\bibliography{example, anthology}
\appendix
\section{Hyperparameters}
\label{sec:hyper}
All the experiments are conducted on 1 GeForce GTX 1080Ti. We set the peak learning rate to 1e-5 and warmup steps to 20 steps. The batch size of ViLBERT COCO and ViLERT HM is set to 16, while that of all the other models is set to 32. We also assign a weight loss of 0.3 to the negative label and 1 to the positive label. They are based on \citet{velioglu2020detecting} with adjustments according to the environment and the characteristics of this dataset.

\section{Details of Dataset Construction}
\label{sec:dataset-detail}
\begin{table*}
\centering
\begin{tabular}{ll}
\toprule
\textbf{Time} & \textbf{Keywords}\\
\midrule
\multirow{5}{*}{August 2021} & vaccine, vaccination, pfizer, moderna, johnson, AstraZeneca, covid, cdc, fauci, \\
&social distancing, wuhan, fda, nih, tedros, 
quarantine, pandemic, bat, sars, \\
& world health organization, vegan, 
vegetarian, peta, animal rights, animal products, \\
&blm, black lives matter, antifa,
wfh, work from home, afghan, afghanistan, taliban, \\&kabul \\
\midrule
\multirow{5}{*}{January 2022} & vaccine, vaccination, pfizer, moderna, johnson, AstraZeneca, covid, cdc, fauci, \\
&social distancing, wuhan, fda, nih, tedros, 
quarantine, pandemic, bat, sars, \\
& world health organization, vegan, 
vegetarian, peta, animal rights, animal products, \\
&blm, black lives matter, antifa,
wfh, work from home, delta, omicron, variant, \\
&booster, shot, dose, mask, anti-vaxxer \\
\bottomrule
\end{tabular}
\caption{\label{kw}
Keywords used in the two rounds of image retrieval from Reddit.
}
\end{table*}

We conducted two rounds of collection. The first collection was in August 2021, followed by a second one subsequently in January 2022 with adjustments on the keywords as shown in Table~\ref{kw}. The retrieval was conducted in the same sets of subreddits: meme, memes, dankmemes, Memes\_Of\_The\_Dank, MemeEconomy, okbuddyretard, MemesIRL, TheRightCantMeme, ConservativeMemes, ShitLiberalsSay, ConspiracyMemes, forwardsfromgrandma, libertarianmeme, AdviceAnimals, EnoughLibertarianSpam, MockTheAltRight, ComedyCemetery, terriblefacebookmemes, PoliticalCompassMemes, ENLIGHTENEDCENTRISM, ABoringDystopia, and COMPLETEANARCHY. The first collection included another topic: Afghanistan, but the topic was subsequently dropped due to the potential bias embodied in the data collected. The supervised data were only sampled from the data from the first collection.
\begin{figure}
\includegraphics[width=\textwidth]{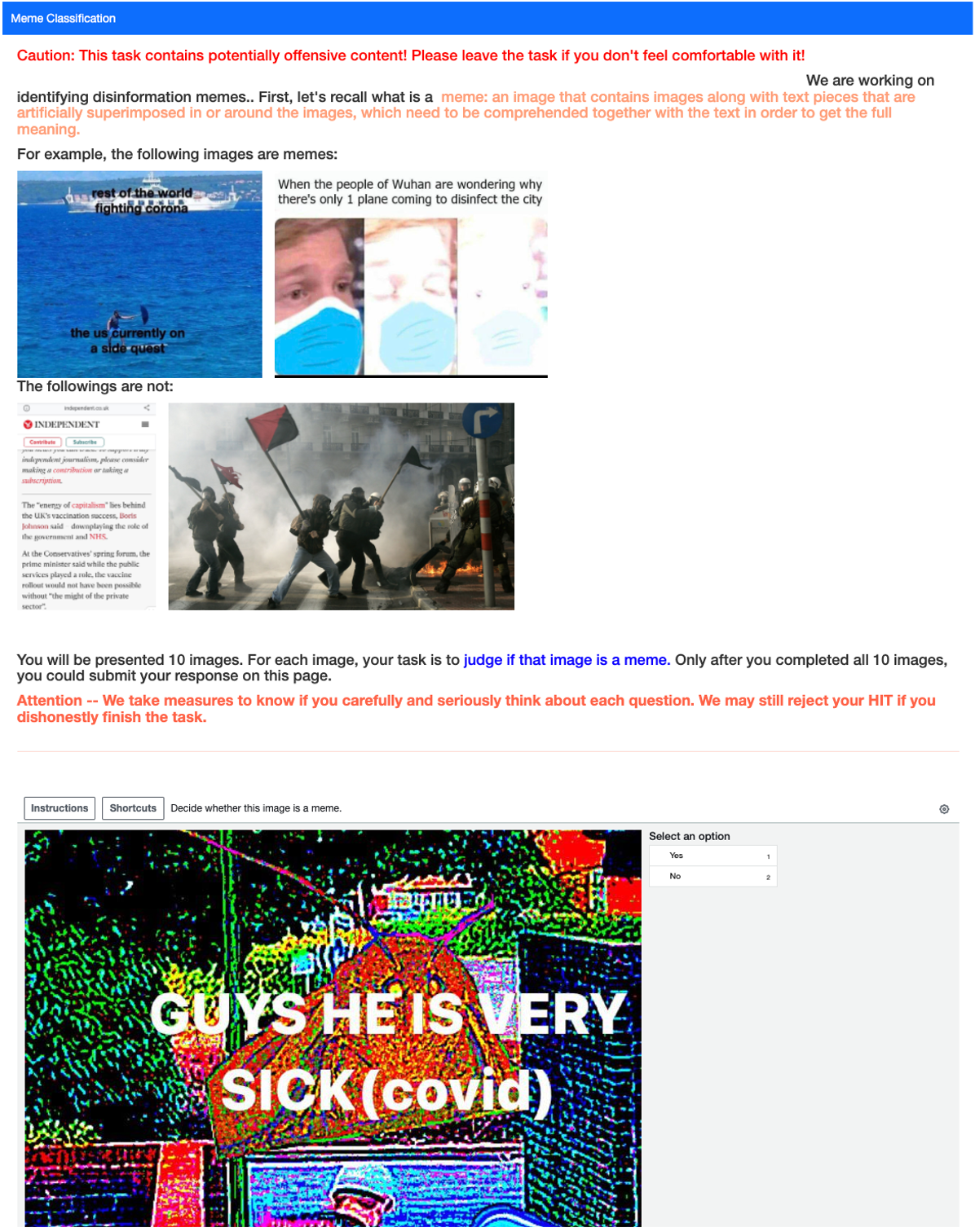}
\caption{\label{Q1}
First question in the annotation process.}
\end{figure}
\begin{figure}
\includegraphics[width=\textwidth]{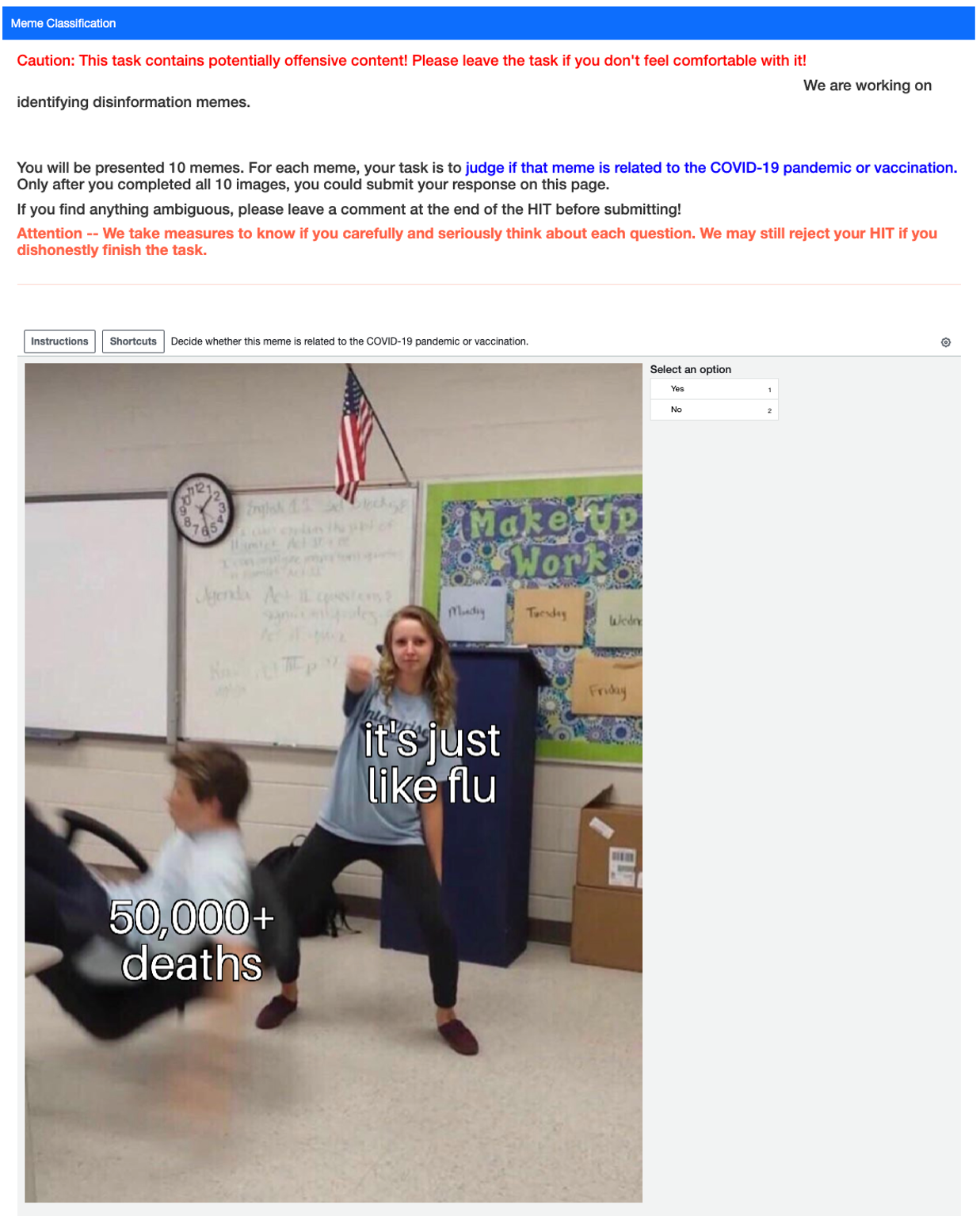}
\caption{\label{Q2}
Second question in the annotation process.}

\end{figure}
\begin{figure}
\includegraphics[width=\textwidth]{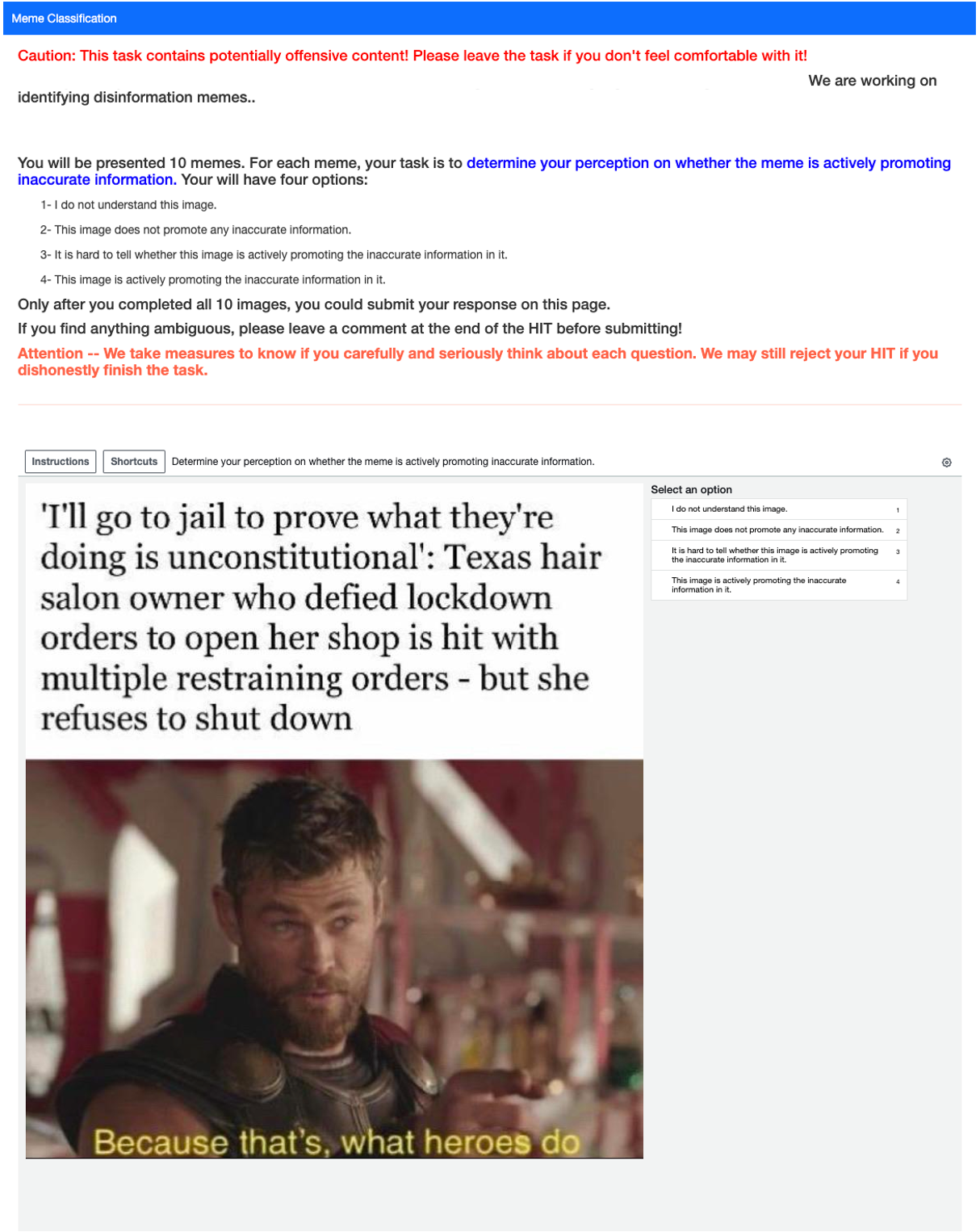}
\caption{\label{Q3}
Third question in the annotation process.}

\end{figure}

Annotation was conducted in two stages. The first stage was a pilot study on 1,000 examples across all topics in October 2021 to test the annotation process. Then, the annotation process was modified to the current process illustrated in Figure~\ref{flow}, and the second annotation of 2,000 examples on Pandemic examples followed in March 2022. As the distribution of labels under the same topic was similar in the two collections (24.90\% positive for the first collection, and 21.55\% positive for the second), we merged the two datasets in our collection by selecting the images from the first stage with an approach that imitates the final annotation process. The heading and an example of an image in each assignment that MTurk workers saw in the current annotation process are shown in Figures~\ref{Q1},~\ref{Q2}, and~\ref{Q3}. To ensure the fairness of payment, one of the authors tested the average time used for each assignment and determined the reward based on a standard of 12 dollars per hour.

\end{document}